\documentclass[letterpaper]{article} 
\usepackage{aaai24}  
\usepackage{times}  
\usepackage{helvet}  
\usepackage{courier}  
\usepackage[hyphens]{url}  
\usepackage{graphicx} 
\usepackage{natbib}  
\usepackage{caption} 
\usepackage{algorithm}
\usepackage{algorithmic}

\usepackage{amsmath}
\usepackage{amssymb}
\usepackage{booktabs}
\usepackage{multirow}
\usepackage{color}

\usepackage{newfloat}
\usepackage{listings}
\DeclareCaptionStyle{ruled}{labelfont=normalfont,labelsep=colon,strut=off} 
\floatstyle{ruled}
\newfloat{listing}{tb}{lst}{}
\floatname{listing}{Listing}
\title{Explicit Visual Prompts for Visual Object Tracking}
\author{
    Liangtao Shi\textsuperscript{\rm 1,\rm 2}, Bineng Zhong\textsuperscript{\rm 1,\rm 2}\thanks{Corresponding author.}, Qihua Liang\textsuperscript{\rm 1,\rm 2}, Ning Li\textsuperscript{\rm 1,\rm 2}, Shengping Zhang\textsuperscript{\rm 3}, Xianxian Li\textsuperscript{\rm 1,\rm 2}
}

\affiliations{
    \textsuperscript{\rm 1} Key Laboratory of Education Blockchain and Intelligent Technology, Ministry of Education, Guangxi Normal University\\
    \textsuperscript{\rm 2} Guangxi Key Lab of Multi-Source Information Mining \& Security, Guangxi Normal University\\
    \textsuperscript{\rm 3} Harbin Institute of Technology\\

    slt@stu.gxnu.edu.cn, bnzhong@gxnu.edu.cn, qhliang@gxnu.edu.cn\\
    ningli65536@mailbox.gxnu.edu.cn, s.zhang@hit.edu.cn, lixx@gxnu.edu.cn\\
}

\usepackage{bibentry}

\begin{document}

\maketitle

\begin{abstract}
How to effectively exploit spatio-temporal information is crucial to capture target appearance changes in visual tracking. However, most deep learning-based trackers mainly focus on designing a complicated appearance model or template updating strategy, while lacking the exploitation of context between consecutive frames and thus entailing the \textit{when-and-how-to-update} dilemma. To address these issues, we propose a novel explicit visual prompts framework for visual tracking, dubbed \textbf{EVPTrack}. Specifically, we utilize spatio-temporal tokens to propagate information between consecutive frames without focusing on updating templates. As a result, we cannot only alleviate the challenge of \textit{when-to-update}, but also avoid the hyper-parameters associated with updating strategies. Then, we utilize the spatio-temporal tokens to generate explicit visual prompts that facilitate inference in the current frame. The prompts are fed into a transformer encoder together with the image tokens without additional processing.
Consequently, the efficiency of our model is improved by avoiding \textit{how-to-update}.
In addition, we consider multi-scale information as explicit visual prompts, providing multiscale template features to enhance the EVPTrack's ability to handle target scale changes. Extensive experimental results on six benchmarks (i.e., LaSOT, LaSOT\rm $_{ext}$, GOT-10k, UAV123, TrackingNet, and TNL2K.) validate that our EVPTrack can achieve competitive performance at a real-time speed by effectively exploiting both spatio-temporal and multi-scale information. Code and models are available at https://github.com/GXNU-ZhongLab/EVPTrack.
\end{abstract}

\section{Introduction}
\begin{figure}[t]
  \centering
   \includegraphics[width=0.9\linewidth]{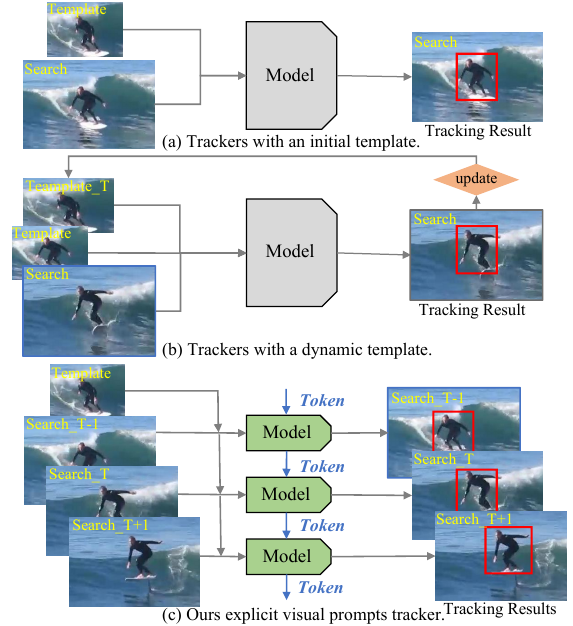}
   \caption{Comparison of tracking frameworks. (a) The framework with an initial template\cite{SiamFC,transt}. (b) The framework with a dynamic template\cite{stark,mixformer}. (c) Our EVPTrack framework uses tokens to propagate spatio-temporal information.}
   \label{figure:aptrack}
\end{figure}

Visual object tracking is a fundamental task in computer vision that aims to locate targets in subsequent frames of a video given an arbitrary target in the first frame. It plays an important role in traffic flow monitoring, human-computer interaction, surveillance, security, etc. Although various top-performing trackers have been proposed over the years, it is still challenging due to target appearance changes, deformation, occlusions,  background clutter, and other factors.

In recent years, deep learning-based trackers \cite{SiamFC, SiamRPN, SiamRPN++, Siamban, transt, ostrack, seqtrack} have achieved excellent performance and efficiency with their powerful representation capabilities, leading the rapid development of the tracking field. Generally speaking, as shown in 
Fig. \ref{figure:aptrack}, they can be divided into two categories: \textit{the trackers with an initial fixed template} and \textit{the trackers with a dynamic template}. The representative trackers with an initial fixed template include siamese-based trackers \cite{SiamFC, SiamRPN, SiamRPN++, Siamban} and transformer-based trackers \cite{transt, ostrack, seqtrack}. The siamese-based trackers formulate the tracking task as a similarity learning problem, matching the initial state of the target with the search frame image to estimate a new target state. 
The transformer-based trackers introduce attention mechanisms\cite{attention} that are utilized for feature extraction or feature fusion. As visualized in Fig. \ref{figure:aptrack}(a), these methods typically use image pairs as input and utilize only the information from an initial template to predict the target positions in the search frames. As a result, they ignore frame-to-frame associations in a video and thus make it challenging to cope with significant target appearance changes. Therefore, some online updated trackers \cite{updateNet, stark, mixformer} introduce template updating mechanisms to enrich the target information and perceive target appearance changes by selecting the search frame as a dynamic template, as shown in Fig. \ref{figure:aptrack}(b). To ensure timely updating and avoid introducing redundant information, the above mentioned methods usually require careful design of the selection strategy and relevant hyperparameters (e.g., score thresholds that control when to update). In addition, some other methods\cite{STMTrack, transformertrack} use spatio-temporal information by introducing memory networks and online update modules, which usually harms efficiency. Despite their success, the above methods face difficulties in deciding when and how to update, and they lack the exploitation of contextual information between consecutive frames.

To address the aforementioned issues, inspired by the philosophy of prompt learning, we propose a novel explicit visual prompts (i.e., spatio-temporal prompts and multi-scale prompts.) tracking framework (EVPTrack) for visual tracking. As illustrated in Fig. \ref{figure:aptrack}(c), EVPTrack utilizes tokens to propagate information between consecutive frames, and generate prompts with spatio-temporal information via the tokens. Specifically, we design a simple Spatio-Temporal Encoder for obtaining new spatio-temporal tokens to propagate the spatio-temporal information to the next frame. And, we design a Prompt Generator for obtaining sptaio-temporal prompts to refine the spatio-temnporal tokens from the previous frame.
In addition, we divide the template image into multiple scale patches to obtain different fine-grained template features to generate multi-scale prompts that enable the tracker to model multi-scale targets efficiently. These explicit visual prompts are fed into Image-Prompt Encoder along with image tokens for relational modeling. This is a novel approach that we propose that can use both spatio-temporal and multi-scale information. Although inspiration comes from prompt learning, it has essential differences from prompt learning: (i) Tasks are different. Prompt learning is to adapt large models to downstream tasks, while we are exploiting temporal information. (ii) Different technical solutions. Prompt learning is by freezing the large model and adding learnable parameters, which is an implicit prompt, whereas we generate prompts using templates and spatio-temporal tokens, which is an explicit visual prompt. 
In summary, our main contributions are as follows:
\begin{itemize}
\item We present a novel explicit visual prompts tracking framework (EVPTrack) that effectively utilizes both spatio-temporal and multi-scale information, and our explicit visual prompts are made from these two kinds of information.

\item To avoid the \textit{when-and-how-to-update} dilemma associated with template update mechanisms, we introduce a novel propagation mechanism that utilizes tokens to propagate spatio-temporal information between consecutive frames.

\item Our tracker achieves state-of-the-art results on six benchmarks. In particular, our method achieved 72.7\% success score (AUC) on the LaSOT test set.
\end{itemize}

\section{Related Work}

{\bf Tracking methods using an initial template.}
The initial template is indispensable, whether it is the previous siamese-based trackers or the mainstream transformer-based trackers nowadays. Generally, the initial template is the target state given in the first frame with reliable apparent information. Therefore, benefiting from the powerful representation capability of deep learning, the template-search based matching paradigm is widely popular due to its simplicity and effectiveness. For example, SiamFC\cite{SiamFC} is a pioneering work in the siamese series of trackers that uses a siamese-network of shared weights to extract search features and template features. Subsequently, it uses the features to perform cross-correlation operations to localize a target in the current frame. To utilize richer semantic information and avoid falling into local optimization due to cross-correlation, transformer-based trackers\cite{transt,stark} utilize the transformer to fusion search features and template features. Recent one-stream trackers\cite{ostrack,simtrack}, which use a transformer-based network to perform both feature extraction and feature fusion, allow for a more full fusion of search features with template features. The above methods can reach a competitive performance by relying only on the initial template frame. However, due to the lack of sensing target changes, they are not robust when facing complex scenes (scale variation, aspect ratio change, etc.) and long-term tracking. Therefore, it is necessary to enable trackers to perceive target changes by exploiting spatio-temporal information.

\begin{figure*}[t]
  \centering
   \includegraphics[width=0.9\linewidth]{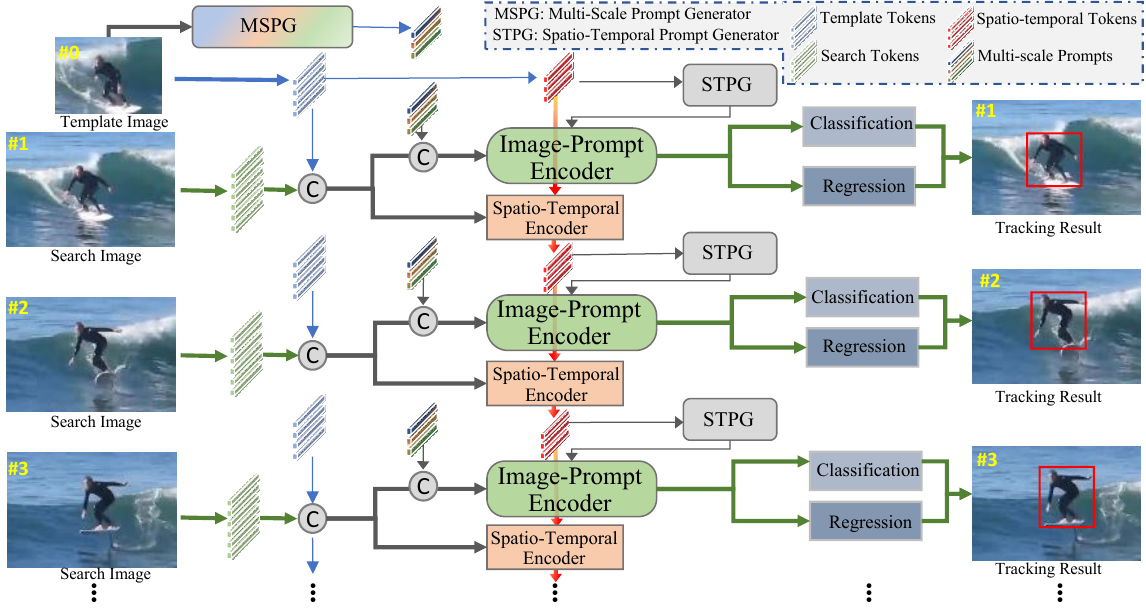}
   \caption{Overview of our framework. The input images are patch embedding to get tokens. Then, Image-Prompt Encoder is used for feature fusion between image tokens and prompts. Finally, the fused search tokens will be used to estimate the target state. In addition, Spatio-Temporal Encoder is used to propagate spatio-temporal information between consecutive frames. Prompt Generator is used to generate explicit visual prompts.
   }
   \label{figure:overview}
\end{figure*}

{\bf Tracking methods using a dynamic template.}
To overcome the above challenges, on the one hand, the DCF-based trackers\cite{atom,PrDiMP-50} try to exploit the temporal information by updating the model online, but they suffer from complex manual optimization. STMTrack\cite{STMTrack} introduces a memory network, which stores the historical information of the target to guide the tracker to focus on the most informative regions in the current frame. TCTrack\cite{TCTrack} exploits consecutive contextual information through online temporal adaptive convolution and adaptive temporal transformers. On the other hand,  template update mechanisms are usually introduced in the siamese-based trackers\cite{updateNet} and the transformer-based trackers\cite{stark,mixformer}. UpdateNet\cite{updateNet} designs a CNN network to generate the optimal template available for the next frame to realize the template update function. Stark\cite{stark} designs a target quality branch that evaluates the quality of the target in the current frame to update the dynamic template selectively. Mixformer\cite{mixformer} selects high-quality templates by presenting a valid target prediction score module, similar to STARK. Despite the success of the above methods, there are the following limitations: (1) It requires a complex design of update modules and update strategies to ensure timely updates and avoid introducing redundant information. (2) It cannot fully utilize the context information between consecutive frames. Therefore, in this work, EVPTrack uses tokens to propagate spatio-temporal information that avoids complex update strategies.


\section{Method}
In this section, we will present the proposed EVPTrack method in detail. First, we will briefly describe an overview tracking framework of EVPTrack. Then, we will give an account of each module of the whole framework one by one.  Finally, we will introduce the training and inference pipelines.

\subsection{Overview}
As shown in Fig.\ref{figure:overview}, EVPTrack is a simple end-to-end tracker with Image-Prompt Encoder, Spatio-Temporal Encoder, and Prompt Generator. The Image-Prompt Encoder is used for the fusion of explicit visual prompts with image features to efficiently utilize spatio-temporal and multi-scale information. The Spatio-Temporal Encoder is utilized for the interaction among template tokens, search tokens, and spatio-temporal tokens to propagate spatio-temporal information. The Prompt Generator extracts information from template and spatio-temporal tokens to generate explicit visual prompts (i.e., multi-scale prompts and spatio-temporal prompts). EVPTrack employs image pairs and spatio-temporal tokens as input. Specifically, here the tokens come from the previous frame. One of the images is a template image $z \in \mathbb{R}^{3 \times H_z \times W_z}$. The image is from the target region in the first frame of the video and contains information about the initial appearance of the target and some of the background information. The other image is a search image $x \in \mathbb{R}^{3 \times H_x \times W_x}$. It comes from the region where the target may appear in the subsequent sequences of the video. To summarize, we utilize spatio-temporal and multi-scale information by providing explicit visual prompts to improve the robustness of the tracker in complex scenarios such as target appearance changes and deformation.


\subsection{Image-Prompt Encoder}
Image-Prompt Encoder inputs include search tokens, template tokens, and explicit visual prompts. We use explicit visual prompts to guide feature extraction and fusion without other complex operations. This way of utilizing spatio-temporal information is simple and avoids additional elaborate updated modules. Specifically, we first perform linear projection on the image pairs. Here, we utilize hierarchical patch embedding with a total downsampling stride of 16, and obtain search tokens and template tokens denoted as $f_{z} \in \mathbb{R}^{N_z \times D} $ and $f_{x} \in \mathbb{R}^{N_x \times D}$, respectively. Here, $N_x = H_xW_x/16^2$, $N_z = H_zW_z/16^2$, $D=512$. And we add learnable position encoding ($P_z \in \mathbb{R}^{N_z \times D}$, $P_x \in \mathbb{R}^{N_x \times D}$) to preserve spatial information.
Then, we concatenate $f_{z}, f_{x}, f_{p}$, and feed them into N-layer encoder. Finally, the search features are utilized to locate the target. This process can be described by the following equations: 
\begin{equation}
\begin{split}
&f_{mp}^0 = concat(f_{p}, f_{z}, f_{x}), \\
&f_{mp}^n = encoder(f_{mp}^{n-1}), n=1...N, \\
&f_{mp}^N = LN(f_{mp}^N),
  \label{eq:eq2}
\end{split}
\end{equation}
here, $f_{p}$, $LN$ denote explicit visual prompts and layer normalization, respectively. For a more detailed design of encoder and patch embedding refer to HiViT\cite{hivit}.

\begin{figure}[t]
  \centering
   \includegraphics[width=0.9\linewidth]{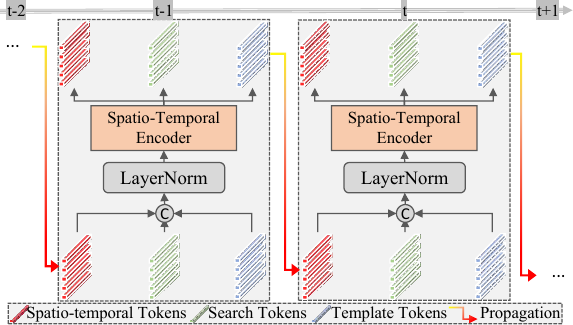}
   \caption{Illustration of Spatio-Temporal Encoder propagation of temporal information.}
   \label{figure:spencoder}
\end{figure}

\subsection{Spatio-Temporal Encoder}
Spatio-Temporal Encoder is a critical component of our framework for maintaining spatio-temporal tokens and thus exploiting the rich contextual information in successive frames. Its structure is shown in Fig.\ref{figure:spencoder}, which is also based on the transformer encoder implementation. Specifically, it uses template, search, and spatio-temporal tokens as input, and obtains new spatio-temporal tokens after feature fusion by transformer encoder. This process can be described by the following equations: 
\begin{equation}
\begin{split}
&f_{in} = concat(f_{t-1}, f_{z}, f_{x}), \\
&f_{out} = encoder(LN(f_{in})), \\
&f_{t} = split(f_{out})[f_{z}], 
  \label{eq:eq3}
\end{split}
\end{equation}
here, $f_{t-1}$ denotes the spatio-temporal tokens obtained at time t-1, which will be used at moment t. When t = 1, we initialize it using template tokens, i.e., $f_{0} = f_{z}$, which ensures that the initialized spatio-temporal tokens can be adapted to any video sequence and can enhance the target features. Then, we split the template tokens into $f_{out}$ as new spatio-temporal tokens. This is to avoid heavy accumulation of mistakes in the spatio-temporal tokens propagated during long-term tracking.

We propagate information between successive frames via tokens, thus effectively exploiting spatio-temporal information to capture the target appearance changes. In contrast to intermittent template updating mechanisms, we utilize a mechanism that allows for continuous propagation of spatio-temporal information. Therefore, avoiding the difficult problem of when-to-update that appears with intermittent template updates.

\subsection{Prompt Generator}
\begin{figure}[t]
  \centering
   \includegraphics[width=0.9\linewidth]{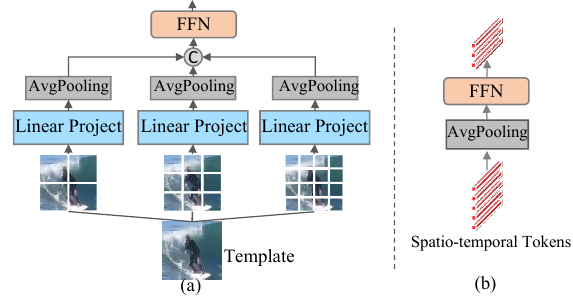}
   \caption{(a): Illustration of multi-scale prompt generator. (b): Illustration of spatio-temporal prompt generator.}
   \label{figure:promptg}
\end{figure}

EVPTrack is equipped with prompt generators to obtain explicit visual prompts (i.e., spatio-temporal prompts and multi-scale prompts). On the one hand, the role of spatio-temporal prompts is to enable the tracker to capture target appearance changes. On the other hand, the multi-scale prompts provide different fine-grained features, thus maintaining excellent tracking even when variations in target scale. For multi-scale prompt generator, whose structure is shown in Fig.\ref{figure:promptg}(a). We first crop the template image into patches with resolution ($P, P$). Then, each patch is mapped to a 1D vector of size D using a learnable linear projection $E \in \mathbb{R}^{(3 \times P^2) \times D}$. In order to obtain features at different scales, here we use three sizes of resolution, P is set to 14,16,18 respectively. The features $f_{p14},f_{p16},f_{p18}$ are obtained corresponding to three different scales. We perform avgpooling on each feature separately and then concatenate to get $f_{ms}$. Finally, $f_{ms}$ are fed into a fully connected network to get the final multi-scale prompts $P_{ms}$. The following equations can describe this process: 
\begin{equation}
\begin{split}
&f_{ms} = concat(avg(f_{p14}), avg(f_{p16}), avg(f_{p18})), \\
&P_{ms} = FFN(f_{ms}). 
  \label{eq:eq1}
\end{split}
\end{equation}

As shown in Fig.\ref{figure:promptg}(b), the spatio-temporal prompt generator uses spatio-temporal tokens as input. First, the input is average pooled by performing avgpooling operation. Then, the extracted features are fed into a fully connected network to obtain the final prompts. Although, these prompt generators are simple in design, it is shown in the experimental section that the generated explicit visual prompts are effective. Therefore, EVPTrack using a small number of explicit visual prompts reduces the computational burden and avoids introducing too much redundant information.

\subsection{Training and Inference}

\noindent{\bf Offline Training.}
EVPTrack uses a conventional center point prediction\cite{ostrack} that contains a regression branch and a classification branch for estimating the target bounding box's and determining the target box location, respectively. During training optimization, For the classification branch we use Gaussian weights focal loss\cite{focalloss}, For the regression branch, a combined L1 loss and GIoU loss \cite{giouloss} is employed.
The total loss function is denoted as:
\begin{equation}
    \mathcal{L}=\mathcal{L}_{cls} + \lambda_1 \mathcal{L}_1+\lambda_2 \mathcal{L}_{\text {giou }} ,
    \label{eq:loss}
\end{equation}
here, $\lambda_1 = 5$ and $\lambda_2 = 2$ are regularization parameters.

\noindent{\bf Online Inference.}
During the inference process, we used post-processing on the prediction results in order to smooth the target movement and changes. Specifically, similar to trackers\cite{ostrack, transt}, a penalty is applied to the classification scores using the Hanning window. After performing the penalty, the box corresponding to the highest score point is selected thereby obtaining the prediction bounding box.

\section{Experiments}

\subsection{Implementation Details}

Our methods are implemented based on python3.8 and pytorch1.10 framework. Our trackers were trained on 4 NVIDIA A10 GPUs. During the inference phase, the trackers were tested at speed on a single NVIDIA RTX2080Ti.

{\bf Models.}
We train two variants of EVPTrack with different input image pair resolutions as follows:
\begin{itemize}
    \item {\bf EVPTrack-224.} Template size: 112x112 pixels. Search region size: 224x224 pixels.
    \item {\bf EVPTrack-384.} Template size: 192x192 pixels. Search region size: 384x384 pixels.
\end{itemize}


{\bf Training strategy.}
We use HiViT-Base\cite{hivit} model as the Image-Prompt Encoder and its parameters are initialized with MAE\cite{MAE}.
EVPTrack is trained on the same datasets as mainstream trackers\cite{ostrack}, including LaSOT\cite{lasot}, GOT-10k \cite{got10k}, TrackingNet\cite{trackingnet}, COCO \cite{coco}. During training, data augmentation employed horizontal flip and brightness jittering, and the optimizer utilized AdamW\cite{adamw}. The learning rate of backbone is set to $1\times10^{-5}$, the learning rate decay is set to $1\times10^{-4}$, and the learning rate of other parameters is set to $1\times10^{-4}$. A total of 150 epochs of training, and each epoch uses 60k search images. The learning rate decreases by factor after 120 epochs. In addition, for a fair comparison of got-10k, we follow its standard protocol and train out the other two models using only the got10k dataset. The models were trained for 50 epochs and started decaying in the 40th epoch as it was trained with only one dataset.

In order to allow EVPTrack to learn the spatio-temporal information between consecutive frames, the sampling strategy during training is therefore different from the traditional sampling strategy. Specifically, we sample M videos in one iteration, each containing N search frames and one template frame. Therefore the batch size is $N*M$. Then, the N search frames are predicted sequentially in order, enabling the tracker to utilize tokens to propagate spatio-temporal information between consecutive frames. For EVPTrack-224, we set N, M to 4 and 8, respectively, with a batch size of 32. EVPTrack-224 is trained on 4 GPUs, so the total batch size is 128. Due to GPU memory constraints, for EVPTrack-384 we set N, M to 4 and 8, respectively.

\begin{table*}[t]
  \centering
  \resizebox{\textwidth}{!}{
     \fontsize{8}{9}\selectfont
  \begin{tabular}{c|c|ccc|ccc|ccc|c}
    \toprule
  \multicolumn{1}{c|}{\multirow{2}{*}{Method}} & \multicolumn{1}{c|}{\multirow{2}{*}{Source}} & \multicolumn{3}{c|}{LaSOT} & \multicolumn{3}{c|}{$\rm LaSOT_{ext}$} & \multicolumn{3}{c|}{GOT-10K$^*$} & \multicolumn{1}{c}{UAV123}\\ \cline{3-12}
    && AUC & $\rm P_{norm}$ & P & AUC & $\rm P_{norm}$ & P  & AO & $\rm SR_{0.5}$ & $\rm SR_{0.75}$ & AUC \\
    \midrule
    
    SiamFC \cite{SiamFC} & ECCVW16 & 33.6 & 42.0 & 33.9 &23.0 &31.1 & 26.9& 34.8 & 35.3 & 9.8 & 46.8\\
    MDNet \cite{MDNET} & CVPR16 & 39.7 & 46.0 & 37.3 & 27.9& 34.9& 31.8& 29.9  & 30.3  & 9.9 & 52.8\\
    ECO \cite{ECO} & ICCV 17 & 32.4  & 33.8 & 30.1  &22.0 & 25.2& 24.0& 31.6  & 30.9  & 11.1 & 53.5\\
    SiamPRN++ \cite{SiamRPN++} & CVPR19 & 49.6  & 56.9  & 49.1  &34.0 &41.6 &39.6 & 51.7  & 61.6  & 32.5 &61.0\\
    Ocean \cite{Ocean}&  ECCV 20 & 56.0  & 65.1  & 56.6  &-& -&-& 61.1 & 72.1  & 47.3 &-\\
    TrDiMP\cite{TrDiMP} & CVPR21 & 63.9  &  - & 61.4  &- &- & -&67.1 & 77.7 & 58.3 &67.5\\
    TransT \cite{transt}& CVPR21  & 64.9  & 73.8 & 69.0 & -& -& -& 67.1  & 76.8  & 60.9 &69.1\\
    AutoMatch\cite{AutoMatch} & ICCV21  & 58.3 & -  & 59.9  & -& -&- & 65.2  & 76.6  & 54.3 &-\\
    STARK\cite{stark} & ICCV21 & 67.1  & 77.0  & - & -&- & -& 68.8 & 78.1 & 64.1 &-\\
    GTELT\cite{GTELT} &CVPR22 &67.7&-&-  &45.0 & 54.2 & 52.2 &- &-&- &-\\
    AiATrack\cite{aiatrack}  &ECCV22 &69.0 & 79.4 &73.8   &- &-&-  &69.6&80.0&63.2 &70.6\\ 
    
    MixFormer-22k \cite{mixformer} & CVPR22 & 69.2  & 78.7  & 74.7 & - & - & - & 70.7  & 80.0  & 67.8 &70.4\\
    SimTrack-B/16\cite{simtrack} & ECCV22 & 69.3 & 78.5 &-  &- &- &- &68.6&78.9&62.4 &69.8\\ 
    OSTrack-384\cite{ostrack} & ECCV22 & 71.1  & 81.1  & 77.6 & 50.5 & 61.3 & 57.6 & 73.7 & 83.2 & 70.8 &\color{blue}70.7\\

    VideoTrack\cite{VideoTrack} & CVPR23 & 70.2  & - & 76.4 & - & - & - & 72.9 & 81.9 & 69.8 &69.7\\
    
    SeqTrack-B384\cite{seqtrack} & CVPR23 & 71.5  & 81.1  & 77.8 & 50.5& 61.6 & 57.5 & 74.5 & \color{blue}84.3 & 71.8 &68.6\\
    
    ARTrack-384\cite{ARTrack} & CVPR23 & \color{blue}72.6  & \color{blue}81.7  & \color{blue}79.1 & \color{blue}51.9 & \color{blue}62.0 & \color{blue}58.5 &  \color{blue}75.5 & \color{blue}84.3 & \color{red}74.3 &70.5\\
    
    \midrule
    EVPTrack-224 & Ours & 70.4  & 80.9  & 77.2 & 48.7 & 59.5 & 55.1 &73.3 & 83.6 & 70.7 &70.2\\
    EVPTrack-384 & Ours & {\color{red}72.7}  & {\color{red}82.9}  & {\color{red}80.3}  & {\color{red}53.7} & \color{red}{65.5}  & \color{red}{61.9}  &  \color{red}76.6  & \color{red}86.7 & \color{blue}73.9 & \color{red}70.9\\
    \bottomrule
  \end{tabular} }
  \caption{ Performance comparisons with state-of-the-art trackers on the test set of LaSOT, $\rm LaSOT_{ext}$ , GOT-10K and UAV123. We add a symbol * over GOT-10k to indicate that the corresponding models are only trained with the GOT-10k training set. The top two results are highlighted with red and blue fonts, respectively.}
  \label{tab:all}
\end{table*}

\subsection{Comparison with State-of-the-art Trackers}


\begin{figure}[t]
  \centering
   \includegraphics[width=0.8\linewidth]{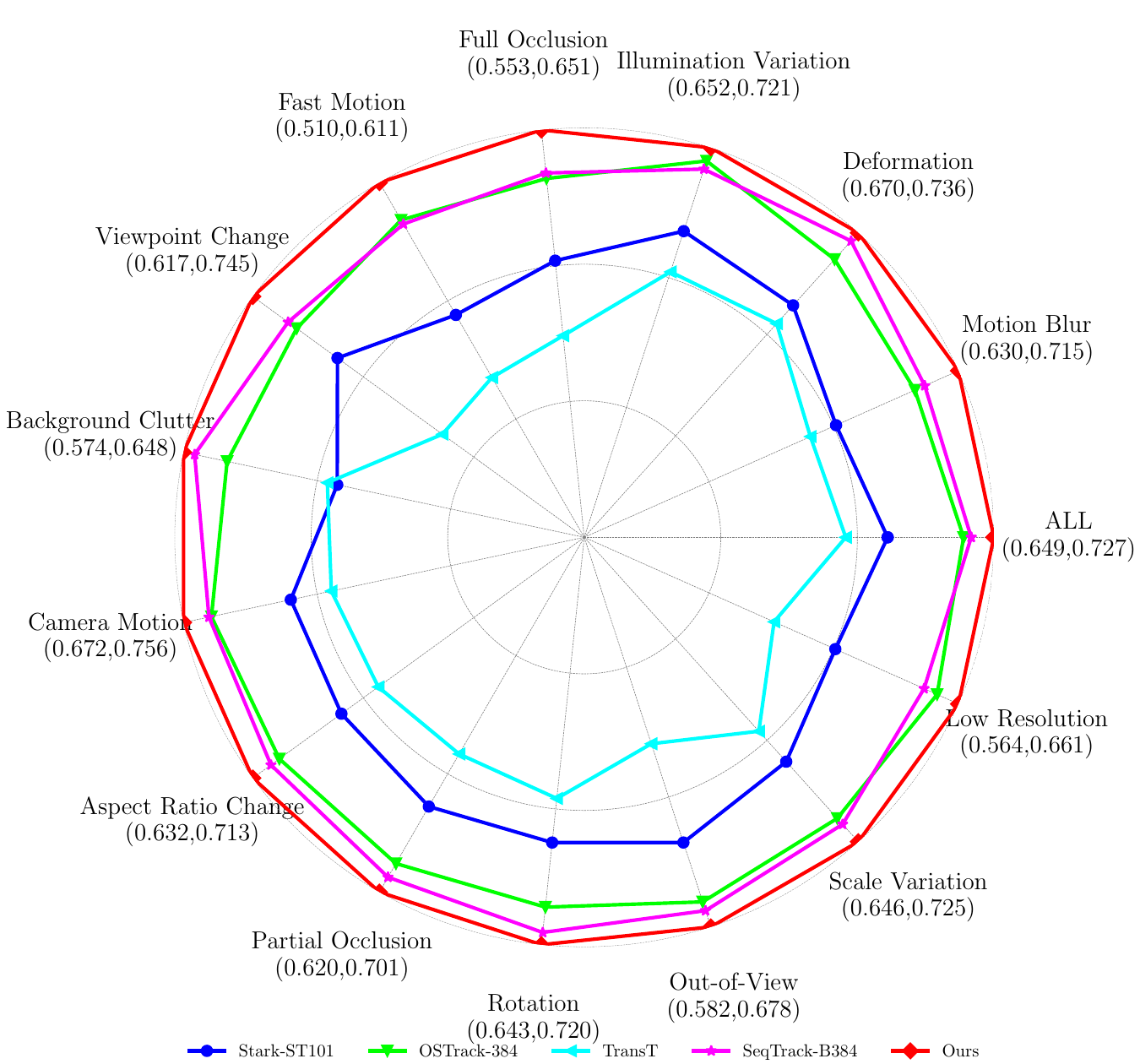}
   \caption{AUC scores of different attributes on LaSOT.}
   \label{figure:lasot}
\end{figure}

{\bf LaSOT.}
LaSOT \cite{lasot} test dataset is a challenging long-term tracking benchmark dataset containing 280 video sequences to effectively evaluate the long-term tracking performance of trackers. EVPTrack's evaluation results are shown in Tab.\ref{tab:all}, EVPTrack outperforms existing trackers by obtaining 72.7\%, 82.9\% and 80.3\% for AUC, P and \rm $P_{norm}$ respectively. In addition, Fig.\ref{figure:lasot} demonstrates the performance of EVPTrack-384 in different challenge scenarios (e.g., scale variation, aspect ratio change). Illustrating that our SeqTrack-B384 performs better than other competing trackers on almost all challenge scenarios. 

{\bf LaSOT$_{ext}$.}
$\rm LaSOT_{ext}$ \cite{lasot-ext} is an expansion of LaSOT with the addition of 150 videos from 15 new categories. There are many similar interfering objects and fast-moving small objects in the added videos, so it is a challenging dataset. As shown in Tab.\ref{tab:all}, although EVPTrack-224 has only 48.7\% AUC, EVPTrack-384 obtains 53.7\% AUC and 61.9\% P at larger input resolutions. EVPTrack-384 outperforms EVPTrack-224 by 5\% while outperforming all comparative trackers. The reason for this disparity, we consider, is that the low resolution lacks enough background context to distinguish between similar objects.

{\bf TrackingNet.}
TrackingNet \cite{trackingnet} is a large-scale tracking benchmark, that contains 511 video sequences, which covers diverse object classes and scenes. It does not provide ground-truth and needs to submit the prediction results to its server to get the Success(AUC) and Precision(P and $P_{norm}$). The returned results are shown in Tab. \ref{tab:trackingnet}, our EVPTrack-384 obtains 84.4\% , 89.1\% and 84.2\% in terms of AUC, P, and $P_{norm}$, respectively. Our EVPTrack-384 can still achieve an advanced performance.

{\bf GOT-10k.}
GOT-10k \cite{got10k} is a large-scale challenging tracking benchmark with over 10k video segments and has 180 segments for the test set. One of the challenges of this dataset is that there is no overlap between the object classes in the training and test sets, except for the general classes of motion objects and motion patterns. We follow the defined protocol proposed in \cite{got10k} and train our model only using the GOT-10k training set and submit the tracking output to the official evaluation server. As reported in Tab.\ref{tab:all}, EVPTrack-384 obtains 76.6\% and 86.7\% in terms of AO and \rm $SR_{0.5}$, respectively. Furthermore, EVPTrack-384 achieved the top-ranked performance on AUC, reaching 76.6\%, surpassing all other trackers. This strongly verifies that our method has strong generalization and is not sensitive to categories.

{\bf TNL2K.}
TNL2K is a recently publicly released large-scale tracking dataset, which contains 3000 challenging video sequences: 700 for testing and 2300 for training. We evaluated our tracker with a test set of 700 video sequences. From Tab. \ref{tab:tnl2k}, we can observe that our tracker achieves a state-of-the-art performance on this dataset. In particular, EVPTrack-224 obtained 57.5\% AUC, even more than SeqTrack-384 using a higher resolution.

{\bf UAV123.}
The UAV123 \cite{uav123} contains 123 video sequences that are captured from the UAV platform. It is a challenging aerial video dataset. The benchmarks can be used to assess whether the tracker is suitable for deployment to a UAV123 in real-time scenarios. From Tab. \ref{tab:all}, we can observe that our tracker performs the best among all compared trackers.

\begin{table*}[t]
\centering
\resizebox{\textwidth}{!}{
\begin{tabular}{c|ccccccccccc|cc}
\toprule
& SiamFC & ECO & SiamPRN++ & ATOM & TransT &STARK & AutoMatch & OSTrack & SimTrack & SeqTrack-B & ARTrack & \textbf{EVPTrack-224} & \textbf{EVPTrack-384} \\
\midrule
AUC & 57.1 & 55.4 & 73.3 & 70.3 & 81.4 & 82.0 & 76.0 & 83.9 & 82.3 & 83.9 & \color{red}85.1 & 83.5 & \color{blue}84.4 \\

$\rm P_{norm}$ & 66.3 & 61.8 & 80.0 & 77.1 & 86.7 & 86.9 & - & 88.5 & 86.5 & \color{blue}88.8 & \color{red}89.1 & 88.3 & \color{red}89.1 \\

\bottomrule
\end{tabular} }
\caption{Comparison with state-of-the-art methods on TrackingNet test set.}
\label{tab:trackingnet}
\end{table*}

\begin{table*}[t]
\centering
\resizebox{\textwidth}{!}{
\begin{tabular}{c|ccccccccccc|cc}
\toprule
& SiamFC & ECO & Ocean & SiamBAN & ATOM & TransT & AutoMatch & OSTrack& SimTrack & SeqTrack-B & ARTrack & \textbf{EVPTrack-224} & \textbf{EVPTrack-384} \\
\midrule
AUC & 29.5 & 32.6 & 38.4 & 41.0 & 40.1 & 50.7 & 47.2 & 55.9 & 54.8 & 56.4 & \color{red}59.8 & 57.5 & \color{blue}59.1 \\

P & 28.6 & -& 37.7 & 41.7 & 39.2 & 51.7 & 43.5 & - & 53.8 & - & - & \color{blue}58.8 & \color{red}62.0 \\

\bottomrule
\end{tabular} }
\caption{Comparison with state-of-the-art methods on TNL2K test set. 
}
\label{tab:tnl2k}
\end{table*}

\begin{table}[t]
    \centering
    \resizebox{\linewidth}{!}{
    \fontsize{8}{9}\selectfont
    \begin{tabular}{c|c|cc|c}
    \toprule
     \multicolumn{1}{c|}{\multirow{2}{*}{\#}}
      &\multicolumn{1}{c|}{\multirow{2}{*}{Method}}
      & \multicolumn{2}{c|}{LaSOT} & \multicolumn{1}{c}{UAV123}  \\ \cline{3-5}
    && AUC & P & AUC \\
      \midrule
      1&baseline & 69.4 & 75.9  & 69.0    \\
      2& + explicit visual prompts& \color{red}70.4 & \color{red}77.2  & \color{red}70.2    \\ \midrule
      3& + learnable tokens & 69.6 & 76.2  & 69.1    \\
      4& + multi-scale prompts& 70.0 & 76.7  & 69.6    \\
      5& + spatio-temporal prompts & 69.7 & 76.5  & 69.3    \\
      6& + using template token & 69.2 & 76.0  & 69.5    \\ \midrule
      7& + single scale (14x14) & 69.6 & 76.3  & 69.7    \\
      8& + single scale (16x16) & 68.9 & 75.5  & 69.4    \\ 
      9& + single scale (18x18) & 69.7 & 76.0  & 69.7    \\
      
    \bottomrule
    \end{tabular} }
    \caption{Ablation studies for explicit visual prompts our tracker in LaSOT and UAV123 benchmark.}
    \label{tab:abl1}
\end{table}

\begin{table}[t]
    \centering
    \resizebox{\linewidth}{!}{
        \fontsize{6}{7}\selectfont
    \begin{tabular}{c|c|c|c|cc|c}
    \toprule
     \multicolumn{1}{c|}{\multirow{2}{*}{\#}}
      &\multicolumn{1}{c|}{\multirow{2}{*}{batchsize}}
      &\multicolumn{1}{c|}{\multirow{2}{*}{M}}
      &\multicolumn{1}{c|}{\multirow{2}{*}{N}}
      &\multicolumn{2}{c|}{LaSOT} & \multicolumn{1}{c}{UAV123}  \\ \cline{5-7}
    &&&& AUC & P & AUC \\
      \midrule
      1&32&8 &4& 69.9 & 76.6  & 69.6    \\
      2&32&4 &8& \color{red}70.4 & \color{red}77.2  & \color{red}70.2    \\
      3&32&2 &16& 69.2 & 76.1  & 69.1    \\
      \bottomrule
    \end{tabular} }
    \caption{Ablation studies the sampling strategy of our tracker in the LaSOT benchmark and UAV123 benchmark. The ``M'' denotes the number of videos used in one iteration, and the ``N" denotes the number of search frames in one iteration, which is also the length of the video sequence.}
    \label{tab:abl2}
\end{table}

\subsection{Ablation Study and Analysis}
We use EVPTrack-224 to validate the effectiveness of our method on the LaSOT and UAV123 benchmarks.

{\bf Baseline $v.s.$ EVPTrack.}
In Tab.\ref{tab:abl1}, we compare the performance of the baseline tracker \#1 using the initial template and ours EVPTrack \#2 using explicit visual prompts. The results show that EVPTrack performance improves with the addition of spatio-temporal and multi-scale explicit visual prompts. While the initial template contained reliable targets, it was unable to cope with target target appearance changes during tracking. In contrast, EVPTrack provides multi-scale target information to enhance the target's features and spatio-temporal information to perceive target appearance changes, which helps the tracker accurately predict the bounding box.


{\bf Study on explicit visual prompts.}
As shown in Tab.\ref{tab:abl1}, \#3 adds the same number of learnable tokens as explicit visual prompts, which is a type of implicit prompt. Although implicit prompts can improve performance, they are slightly lower than EVPTrack. This is because the implicit prompts does not change during the inference phase, and the tracker relies solely on initial template information and lacks exploitation of spatio-temporal information. In contrast, our display prompt is adaptively generated from templates and spatio-temporal tokens.

Moreover, we conducted experiments \#4,\#5,\#6,\#7,\#8 and \#9 to explore the effectiveness of the multi-scale prompts and the spatio-temporal prompts. We can observe that there is a significant improvement on the performance of multi-scale prompts relative to single-scale prompts. In experiment \#4 compared to baseline \#1, this improves the performance of the tracker by providing different fine-grained template features. In experiment \#5, we trained EVPTrack to use only spatio-temporal prompts. This improves the performance of the tracker by exploiting spatio-temporal information between consecutive frames compared to baseline \#1. Comparison of experiments \#5 and \#6 demonstrates the effectiveness of the spatio-temporal encoder. In Tab.\ref{tab:abl1}, the best performance is achieved when both prompts are employed. 

{\bf Video sequence length during training.}
As shown in Tab.\ref{tab:abl2}, to explore the spatio-temporal information between consecutive frames, we performed experiments related to the length of the video sequence during training. We keep the batchsize, the learning rate, and the total amount of data constant when training the trackers. Comparing experiments \#1 and \#2, when the sequence length was changed from 4 to 8, the AUC increased by 0.5\% on LaSOT. This demonstrates that the tracker can obtain spatio-temporal information from consecutive video frames. Furthermore, comparing experiments \#2 and \#3, although the sequence length changed from 8 to 16, the AUC decreased by 1.2\% on LaSOT, which may be because only two videos were used in each iteration during training, leading to a lack of generalization of the trained tracker. So we can notice a limitation of our approach: the limited GPUs memory restricts the length of the training video which hinders our tracker from fully exploiting the spatio-temporal information.

{\bf Speed, FLOPs, and Params.}
As demonstrated in Tab.\ref{tab:fps}, our EVPTrack-224 can run in real-time at over 71fps. Furthermore, the FLOPs of EVPTrack-384 are 2x lower than SeqTrack-B384, and the AUC of EVPTrack-384 on LaSOT is 1.2\% higher than that of SeqTrack-B384. This demonstrates that our method can have the effective benefit of temporal and multi-scale information, as well as a good balance between efficiency and performance.

\begin{table}[t]
    \centering
    \resizebox{\linewidth}{!}{
    \fontsize{8}{9}\selectfont
    \begin{tabular}{l|cccc}
    \toprule
      Model & Params(M) & FLOPs(G) &Speed(FPS) & AUC \\
      \midrule
        SeqTrack-b384 & 89 & 148 & 11 &  71.5 \\ 
        EVPTrack-224  & 73 & 21 & 71 &  70.4 \\
        EVPTrack-384  & 73 & 65 & 28 &  72.7 \\   
        \bottomrule
        \end{tabular} }
        \caption{Comparison about the Speed, FLOPs, Params and Performance on LaSOT.}
        \label{tab:fps}
    \end{table}

\begin{figure}[t]
  \centering
   \includegraphics[width=0.9\linewidth]{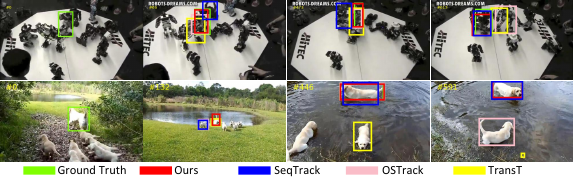}
   \caption{Visualized comparison results of our tracker with three SOTA trackers on LaSOT benchmark.}
   \label{figure:vis}
\end{figure}

{\bf Qualitative comparison.}
In addition, to demonstrate the effectiveness of our method more intuitively, we provide some qualitative comparison results in Fig.\ref{figure:vis}. For example, in the first sequence, compared to the other three SOTA trackers, our EVPTrack can still obtain accurate tracking results in the situation of having similar objects by utilizing explicit visual prompts.

\section{Conclusions}

In this work, we propose a novel explicit visual prompts framework for visual tracking, dubbed EVPTrack. We introduce a novel mechanism of spatio-temporal information propagation which eliminates the need to focus on \textit{when-to-update}. Furthermore, we abstract spatio-temporal and multi-scale information into explicit visual prompts. These prompts are simply fused with the image tokens via transformer encoders without the need to customize additional modules. Therefore, EVPTrack avoids complicated updating strategies. Extensive experiments conducted on six visual tracking benchmarks have shown that the proposed EVPTrack achieves competitive performance at real-time speed, confirming its effectiveness and efficiency.

\section{Acknowledgements}
This work is supported by the National Natural Science Foundation of China (No.U23A20383, 61972167 and U21A20474), the Project of Guangxi Science and Technology (No.2022GXNSFDA035079 and 2023GXNSFDA026003), the Guangxi "Bagui Scholar" Teams for Innovation and Research Project, the Guangxi Collaborative Innovation Center of Multi-source Information Integration and Intelligent Processing, the Guangxi Talent Highland Project of Big Data Intelligence and Application, and the Research Project of Guangxi Normal University (No.2022TD002).

\bibliography{aaai24}

\end{document}